\documentclass[letterpaper, 10 pt, conference]{ieeeconf}  %
\usepackage{siunitx}

\IEEEoverridecommandlockouts                              %

\usepackage[backend=bibtex,bibstyle=ieee,citestyle=numeric-comp,maxbibnames=3,maxcitenames=3,doi=false,url=false,eprint=false]{biblatex}
\usepackage{graphicx}
\usepackage{algpseudocode}
\usepackage{algorithm}
\usepackage{graphics} %
\usepackage{svg}
\usepackage{tabularx}
\usepackage{tablefootnote}

\usepackage{amsmath} %
\usepackage{cleveref}
\usepackage[font=footnotesize]{subcaption}

\DeclareMathOperator*{\argmin}{argmin}

\title{\LARGE \bf
A Chef`s KISS - Utilizing semantic information \\in both ICP and SLAM framework
}

\author{Sven Ochs$^{1}$, Marc Heinrich$^{1}$, Philip Sch\"orner$^{1}$, Marc René Zofka$^{1}$ and J. Marius Z\"ollner$^{1,2}$%
\thanks{$^{1}$ Department of Technical Cognitive Systems, FZI Research Center for Information Technology, 76131 Karlsruhe, Germany.
	{\tt\small ochs, heinrich, schoerner, zofka, zoellner@fzi.de}}%
\thanks{$^{2}$ Karlsruhe Institute of Technology (KIT), Germany.}
 }

\begin{document}

\maketitle
\thispagestyle{empty}
\pagestyle{empty}

\begin{abstract}

For utilizing autonomous vehicle in urban areas a reliable localization is needed. Especially when HD maps are used, a precise and repeatable method has to be chosen. Therefore accurate map generation but also re-localization against these maps is necessary. Due to best 3D reconstruction of the surrounding, LiDAR has become a reliable modality for localization. The latest LiDAR odometry estimation are based on iterative closest point (ICP) approaches, namely KISS-ICP \cite{vizzo_kiss-icp_2023} and SAGE-ICP \cite{cui_sage-icp_2023}. We extend the capabilities of KISS-ICP by incorporating semantic information into the point alignment process using a generalizable approach with minimal parameter tuning. This enhancement allows us to surpass KISS-ICP in terms of absolute trajectory error (ATE), the primary metric for map accuracy. Additionally, we improve the Cartographer mapping framework to handle semantic information. Cartographer facilitates loop closure detection over larger areas, mitigating odometry drift and further enhancing ATE accuracy. By integrating semantic information into the mapping process, we enable the filtering of specific classes, such as parked vehicles, from the resulting map. This filtering improves relocalization quality by addressing temporal changes, such as vehicles being moved.
\end{abstract}

\section{INTRODUCTION}
\label{sec:introduction}

Localization of autonomous vehicles is the first component of most autonomous driving stacks. It allows the vehicle to gain a precise knowledge of the environment. A widespread approach is to encode the information in high definition maps. To find itself in the HD map, a high-precision localization is needed. In open space, an accurate position can be achieved through GNSS solutions or the utilization of real time kinematic data. However, in many situations the GNSS signal is obstructed by trees or buildings. Moreover, in scenarios like parking garages or tunnels there is no signal available. Consequently, other modes of localization are needed. For production use in last mile shuttles like \cite{ochs_stepping_2023} \cite{cornet_novel_2025} \cite{cornet_real-life_2025}, a repeatable result is necessary. Additional challenges for last mile shuttles are the narrow streets in peri urban areas, requiring extra high precision and reproducibility. Among the prominent techniques, visual odometry \cite{lategahn_vision-only_2014}, LiDAR odometry \cite{zhang_loam_2014}, mostly relying on ICP, and multi-sensor fusion odometry \cite{campos_orb-slam3_2021} have garnered significant attention. Cameras, being cost-effective, are extensively utilized in commercial autonomous driving systems; however, their performance can degrade due to factors such as illumination variations and scale ambiguity. LiDAR provides precise distance measurements directly and demonstrates greater robustness to variations in environmental conditions.

\begin{figure}[H]
    \begin{subfigure}[b]{0.49\columnwidth}
		\includegraphics[width=\textwidth]{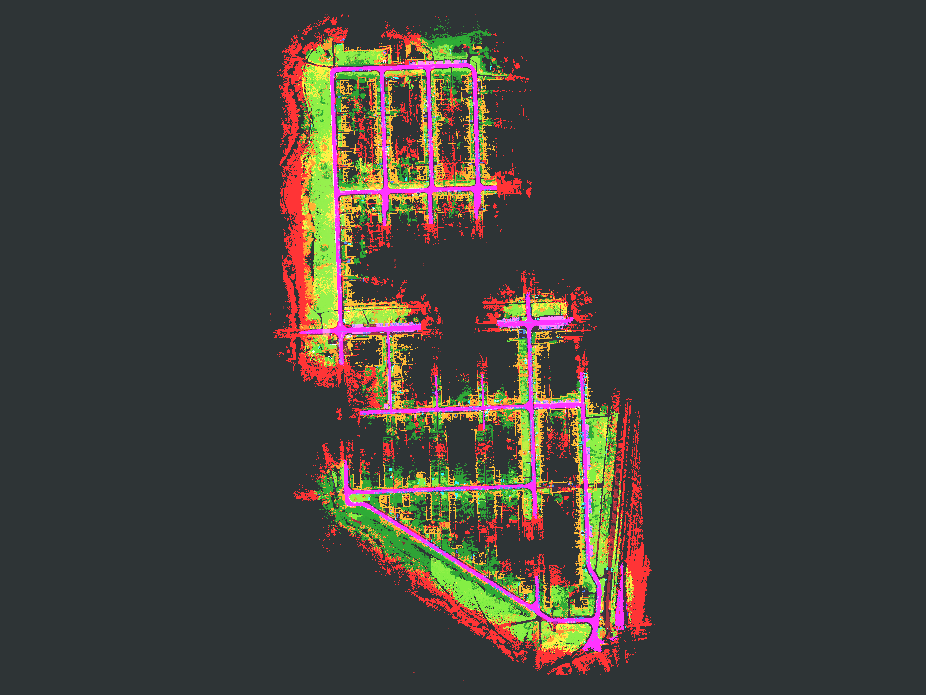}
        \caption{Semantic Map of Scenario 08}
		\label{fig:wd}
    \end{subfigure}
    \begin{subfigure}[b]{0.49\columnwidth}
		\includegraphics[width=\textwidth]{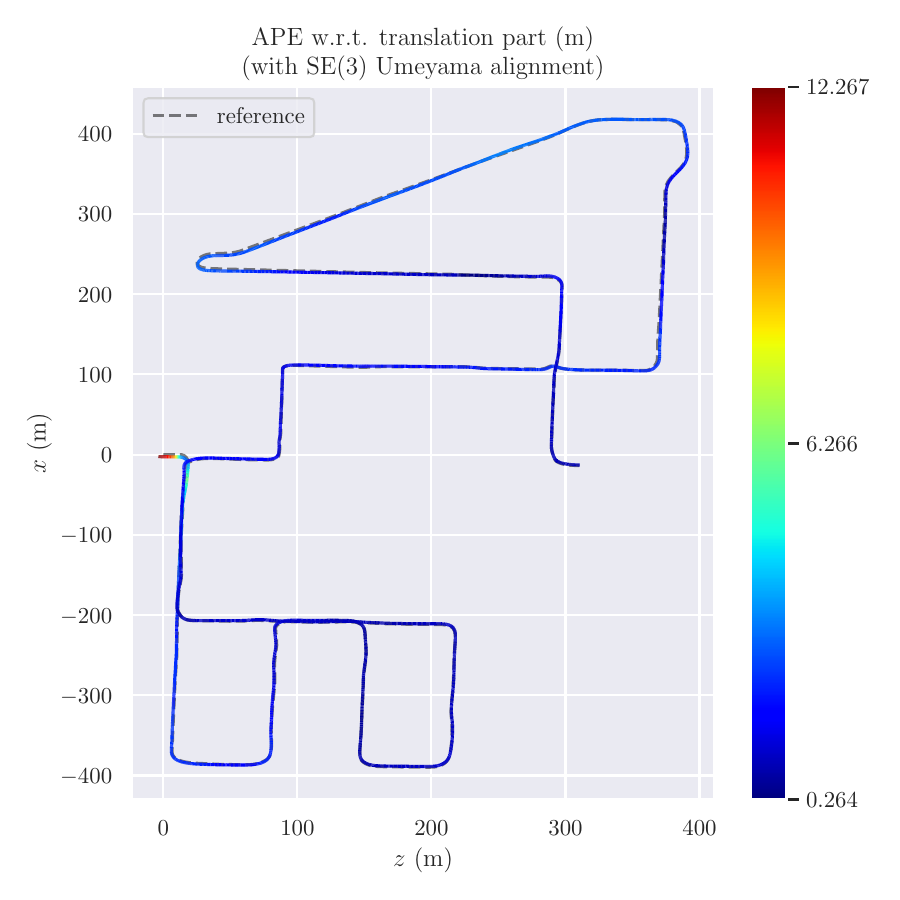}
        \caption{APE-3D of Scenario 08}
		\label{fig:wd-trajectory}
    \end{subfigure}

    \begin{subfigure}[b]{0.49\columnwidth}
		\includegraphics[width=\textwidth]{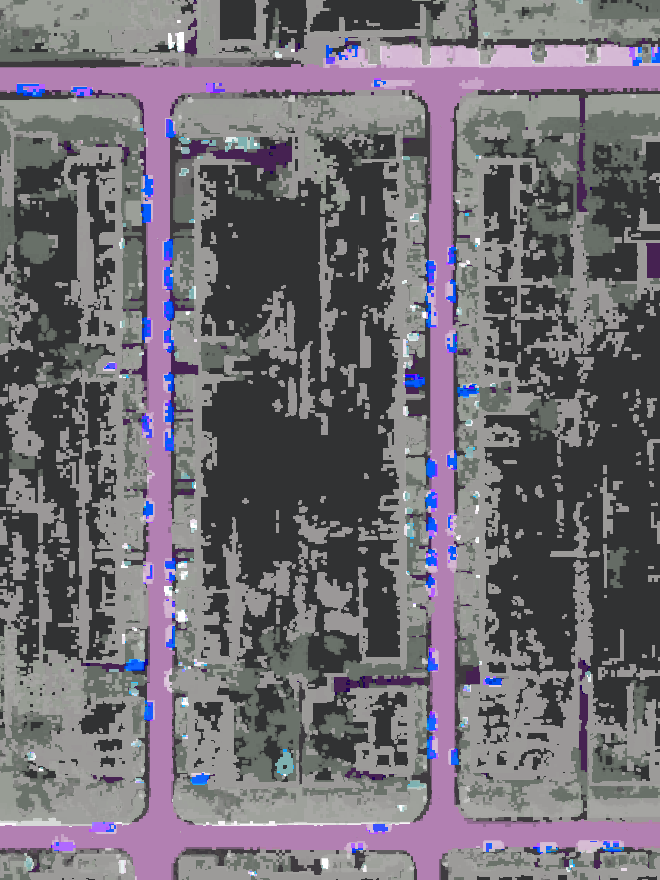}
        \caption{Semantic map with cars}
		\label{fig:wd_zoom}
    \end{subfigure}
    \begin{subfigure}[b]{0.49\columnwidth}
		\includegraphics[width=\textwidth]{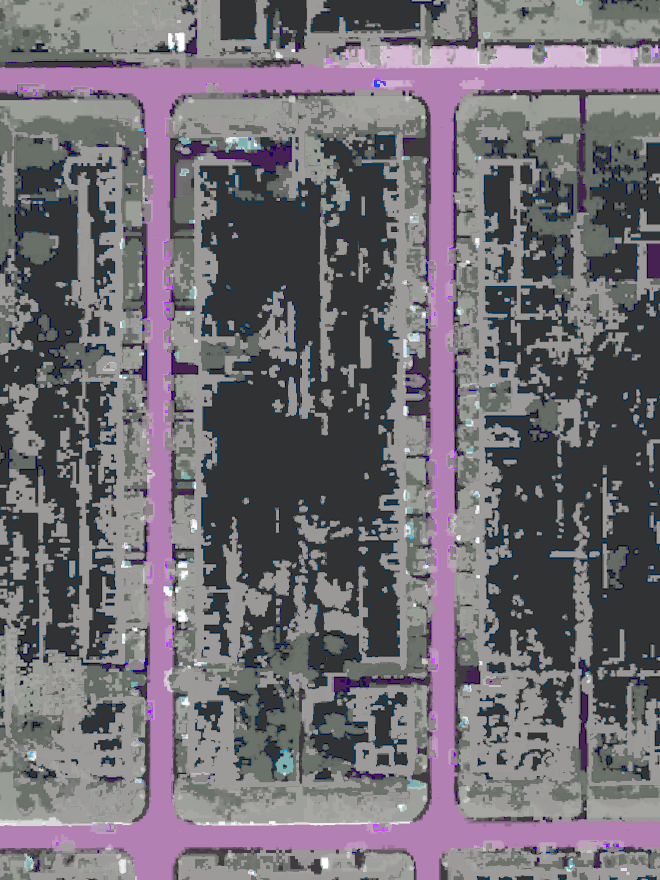}
        \caption{Semantic map withou cars}
		\label{fig:wd_zoom_no-car}
    \end{subfigure}
    
    \caption{The main feature of this paper is the improvement of the Cartographer framework utilizing a semantic adaptation of the KISS-ICP \cite{vizzo_kiss-icp_2023} approach. Due to this reason we outperform the state-of-the-art and also add the capabilities of Cartographer. This includes map manipulation as seen in \cref{fig:wd_zoom} and \cref{fig:wd_zoom_no-car}. The backgrounds are desaturated for better visualization.}
    \label{fig:entery-figure}
\end{figure}

This leads to LiDAR being the preferred mode for localization because it can tackle all the above challenges. It can obtain a precise representation of the environment and can be saved in an arbitrarily defined map. An additional challenge is that the environment is under constant change. This is caused by change of seasons, construction sites, or semi dynamic obstacles like parked vehicles. These semi dynamic obstacles are degrading the localization, since they are static in the mapping run, but during localization runs, may have left or even just moved to a different position. To avoid false localization, this problem can be tackled through additional feature extraction of the LiDAR data. Semantic segmentation provides the classification for the online and offline filtering. The online filtering process eliminates moving objects to prevent potential misalignment caused by their motion during the ICP process. The semi-dynamic obstacles like parked vehicles provide necessary feature for the odometry estimation. These parked vehicle should not be included in the final map, however, due to the previously mentioned point \cite{chen_suma_2019}.

The integration of KISS-ICP\cite{vizzo_kiss-icp_2023}, adapted to semantic labels, with the robust framework capabilities of Cartographer \cite{hess_real-time_2016}, combines the strengths of both approaches. The adapted KISS-ICP demonstrates high accuracy, particularly in reducing absolute trajectory error. This adaptation involves filtering dynamic obstacles, as outlined in \cite{cui_sage-icp_2023}, and modifying the data structure to focus on smaller objects. Furthermore, the weighting of residuals in KISS-ICP has been refined to effectively account for semantic labels. When paired with the loop closure capabilities of Cartographer, the overall performance is further enhanced. Additionally, Cartographer's mapping process has been tailored to accommodate semantically labeled point clouds, enabling offline filtering of the residual map.

\section{RELATED WORK}
\label{sec:related_work}

Dellenbach et al. \cite{dellenbach_ct-icp_2021} provide a iterative closes point (ICP) approach \cite{dellenbach_github_nodate}. CT-ICP uses for scan matching two pose estimations for each scan. One pose is estimated at the start of the 360-degree LiDAR and one at the end. For this reason, they are able to interpolate between those two poses to compensate for high longitudinal velocity and rotational movements. For down sampling the point cloud, a voxel grid is used using a \SI{1}{metre} big voxel. These voxels accept 20 points per voxel. An additional loop closure mechanism based on a elevation map is also introduced.

SuMa++ by Chen et al. \cite{chen_suma_2019} proposed as an extension of a surface element (surfels) based mapping \cite{behley_efficient_2018} approach exploiting three-dimensional laser range scans by integrating semantic information to facilitate the mapping process. The SuMa approach projects the point cloud onto a vertex map, representing a 2D mapping of 3D points. The vertex map is the base for the calculation of the normals, which are then used for an odometry estimation via ICP. The generated map using the vertex maps is then used for a loop-closure search. The result is a semantically enriched map with labeled surface elements, called surfels, in the SuMa++ approach. To allow real-time capable processing even in ever-growing maps, they utilize the GPU for map generation.

Vizzo at al. \cite{vizzo_kiss-icp_2023} propose the KISS-ICP, short for Keep It Small and Simple. The ICP consists of four steps: deskewing utilizing sensor and motion prediction, utilizing a constant velocity model. Followed by down sampling of the point cloud data using a voxel grid based approach. For faster runtime, they use a double down sampling strategy by using the original points without calculating the center point of the voxel. The local map utilizes a hash map to represent the 3D voxel grid. The association is done via a classic point-to-point matching. Due to its simplicity it has only 7 parameters, compared to CT-ICP`s 30 or SUMA`s 49.

Cui et al. introduce SAGE-ICP\cite{cui_sage-icp_2023}, an approach built upon KISS-ICP which is enhanced with semantic information. Their method processes incoming point clouds by filtering out dynamic obstacles and performing semantic sub sampling. The sub sampling process applies varying grid sizes based on the semantic category. For semantic associations, the Euclidean distance is weighted by the semantic distance. The method distinguishes between cases where labels overlap and those where no labels are available. 

LOAM by Zhang and Singh \cite{zhang_loam_2014} and its many derivatives like A-loam, V-Loam, F-Loam \cite{wang_f-loam_2021} and LEGO-Loam \cite{shan_lego-loam_2018} extract planar surface patches and sharp edges from each point cloud. These edge and surface points are then used for finding correspondences over consecutive scans. Through the association of these points an ego motion estimation can be derived. This allows a deskewing using a linear interpolation. The mapping is also utilizing the same algorithm but with 10 times more points. These points are then matched against the map, which is the accumulated point cloud with an applied voxel filter.

Shan et al. propose with lio-sam \cite{shan_lio-sam_2020} an approach that fixes the issues of LOAM, which suffers from drift in large-scale tests, as it is a scan-matching-based method at its core. They deskew the point cloud with a nonlinear motion model, for which they need to use raw IMU measurements. The state estimation is represented as factor graph problem, consisting of a IMU, LiDAR, GPS  and loop closure factors. Through key framing a map is generated, which is utilized for scan matching of incoming frames.

In addition to the previous works, Cartographer\cite{hess_real-time_2016} and MOLA\cite{blanco_claraco_modular_2019} also provide a full ecosystem around the LiDAR-based odometry. They are providing a SLAM algorithm with LiDAR based odometry, but also include the possibility to save the calculated map. These can either be used for post-processing but also for re-localization. In the case of re-localization, the map can be loaded, and the current measurements can be used to determine the position of the mobile system within the map.
MOLA uses a framework that is based on the MRTP and provides a modular framework where multiple components can be exchanged. It is mainly open source especially the underlying LiDAR odometry is available \cite{blanco-claraco_flexible_2024}. But the loop closure module is not public available. In contrast, Cartographer is fully open source. It is not quite as modular as the MOLA framework but also provides the possibility for exchanging the underlying algorithms. The baseline of Cartographer uses ceres to match the incoming point cloud against the submap, which uses probability grids as the underlying data structure. Cartographer is still state of the art in indoor scenarios as in \cite{li_performance_2024}, but in outdoor scenario it is outperformed by the approaches above.

\section{METHOD}
\label{sec:method}

\begin{figure*}[ht]
     \begin{subfigure}[b]{0.245\textwidth}
		\includegraphics[width=\textwidth]{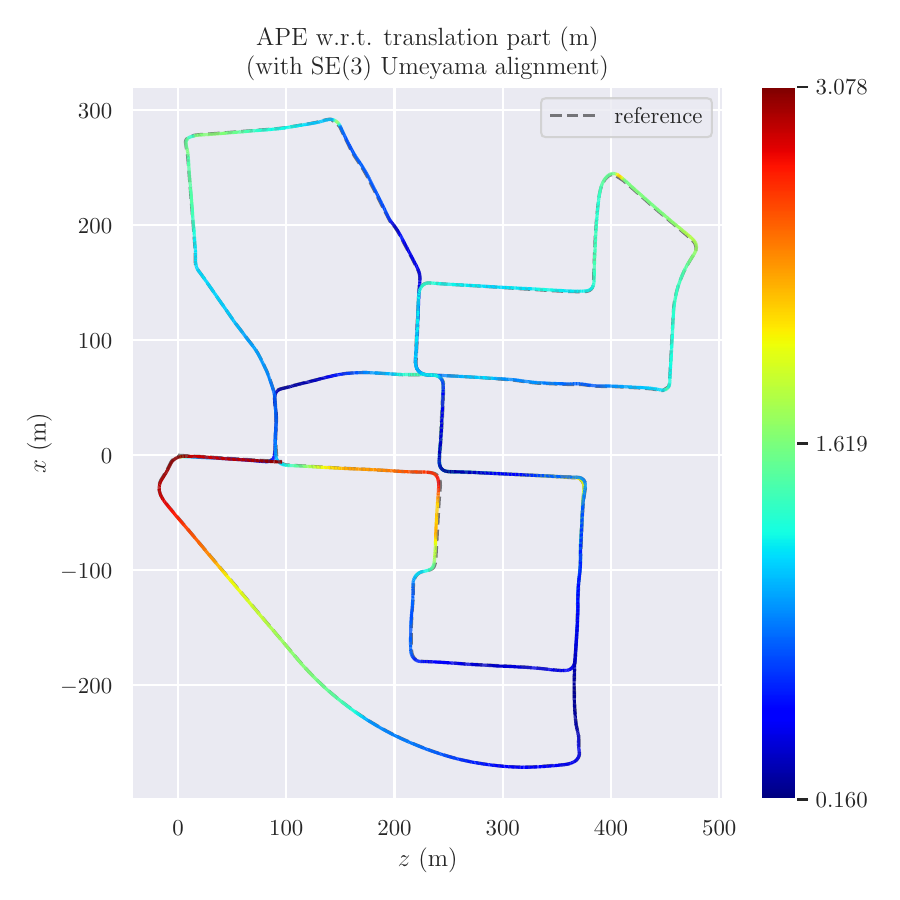}
        \caption{Scenario 00}
		\label{fig:00}
    \end{subfigure}\hfill
    \begin{subfigure}[b]{0.245\textwidth}
		\includegraphics[width=\textwidth]{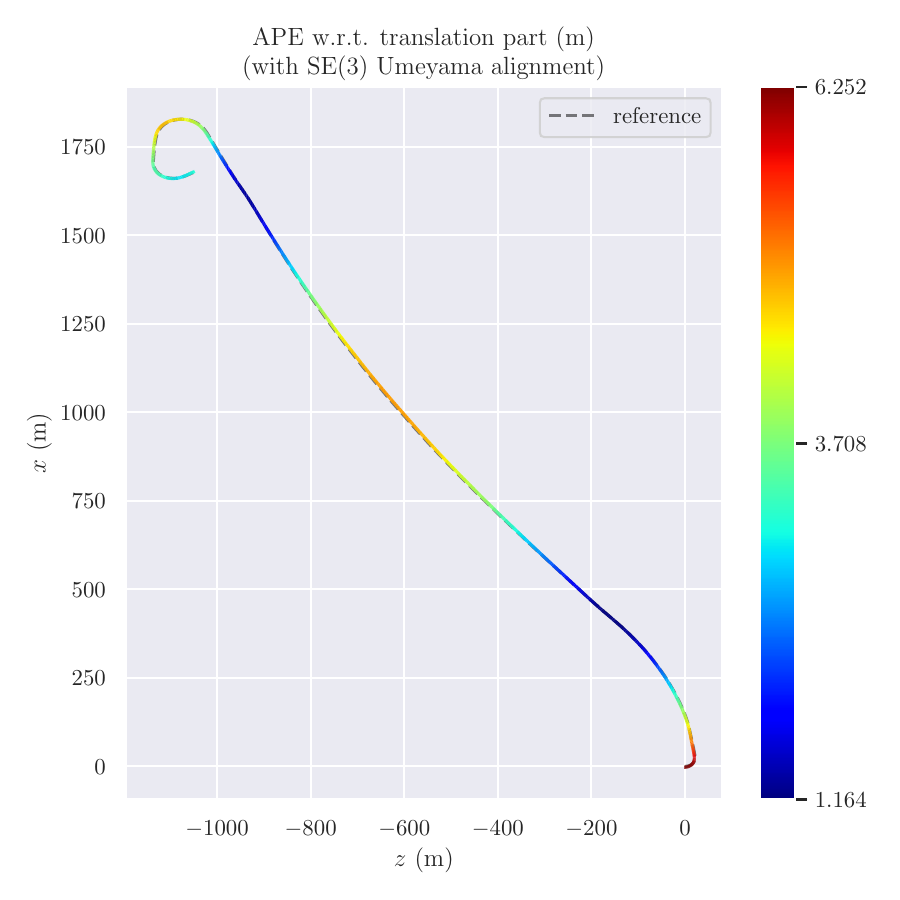}
        \caption{Scenario 01}
		\label{fig:01}
    \end{subfigure}\hfill
    \begin{subfigure}[b]{0.245\textwidth}
		\includegraphics[width=\textwidth]{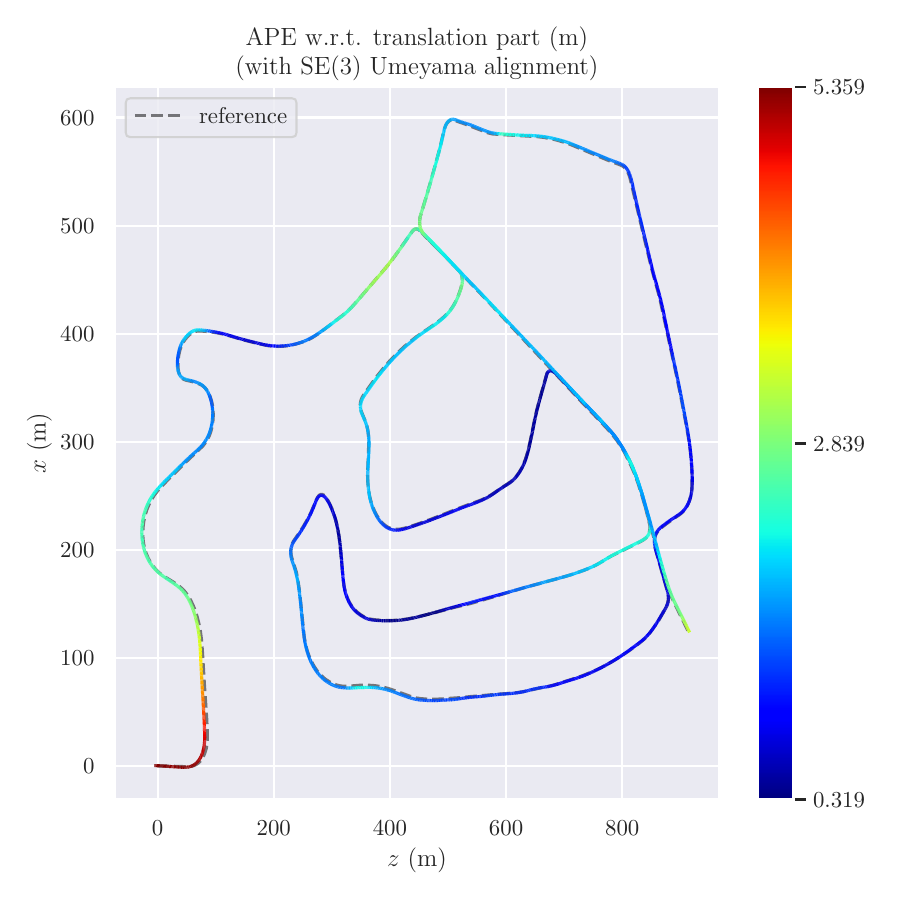}
        \caption{Scenario 02}
		\label{fig:02}
    \end{subfigure}\hfill
    \begin{subfigure}[b]{0.245\textwidth}
		\includegraphics[width=\textwidth]{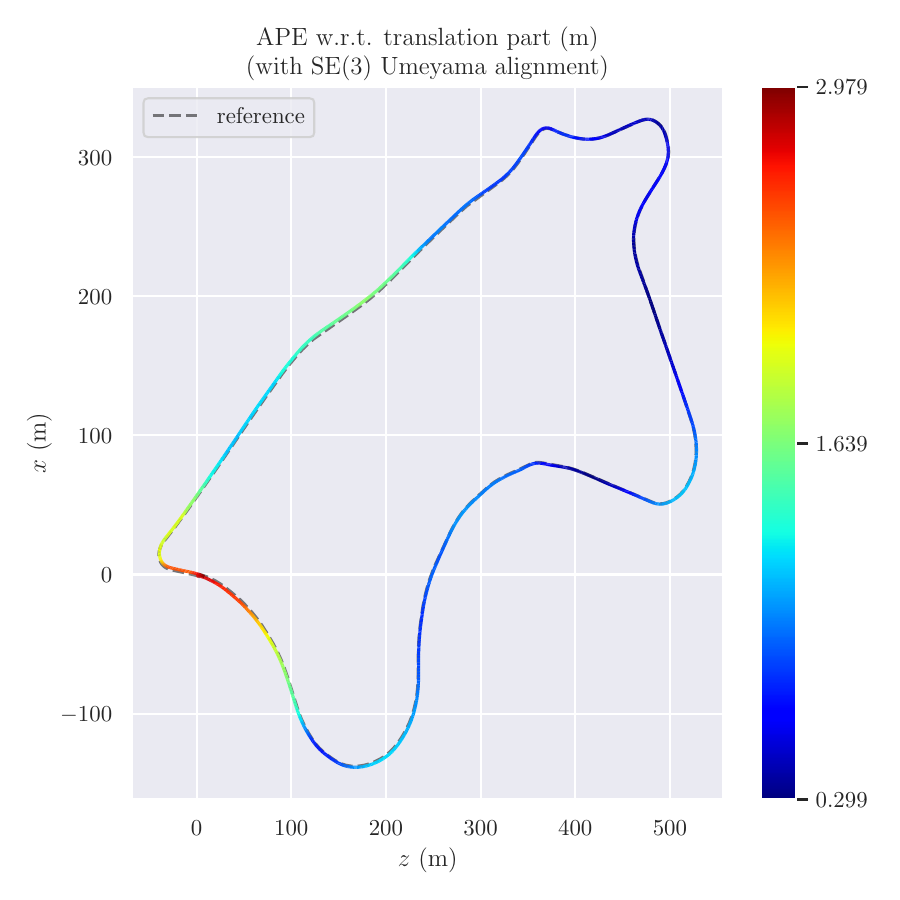}
        \caption{Scenario 09}
		\label{fig:09}
    \end{subfigure}\hfill

    \begin{subfigure}[b]{0.245\textwidth}
		\includegraphics[width=\textwidth]{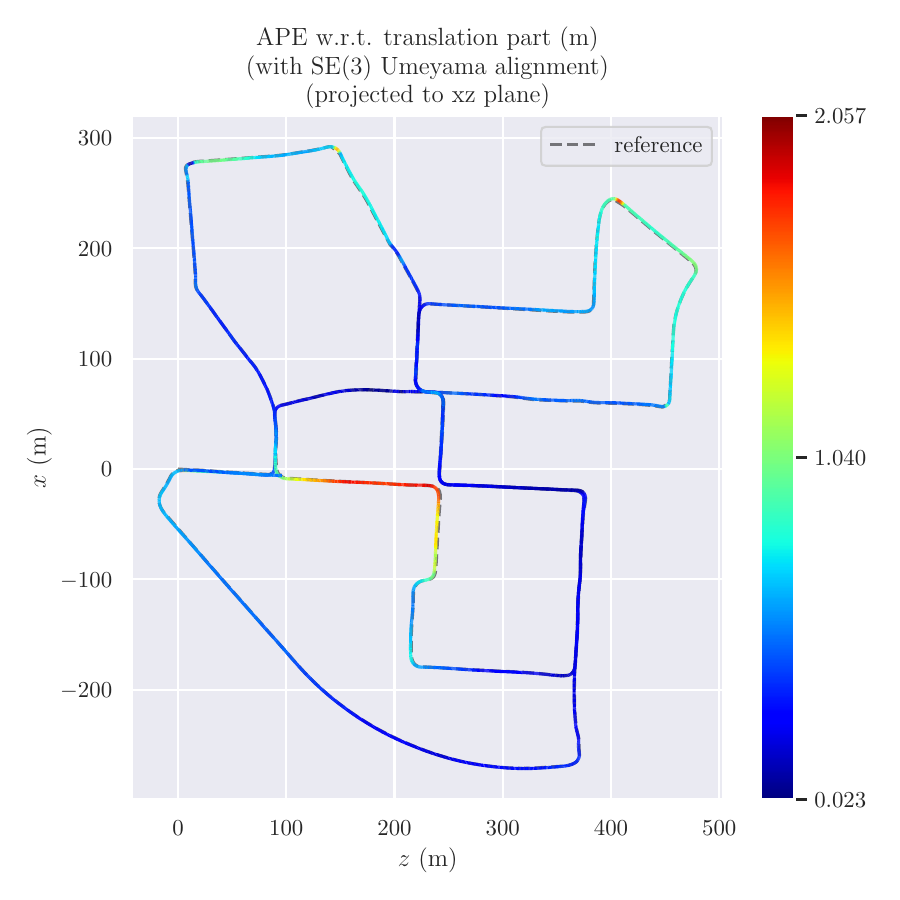}
        \caption{Scenario 00 (2D)}
		\label{fig:00-2d}
    \end{subfigure}\hfill
    \begin{subfigure}[b]{0.245\textwidth}
		\includegraphics[width=\textwidth]{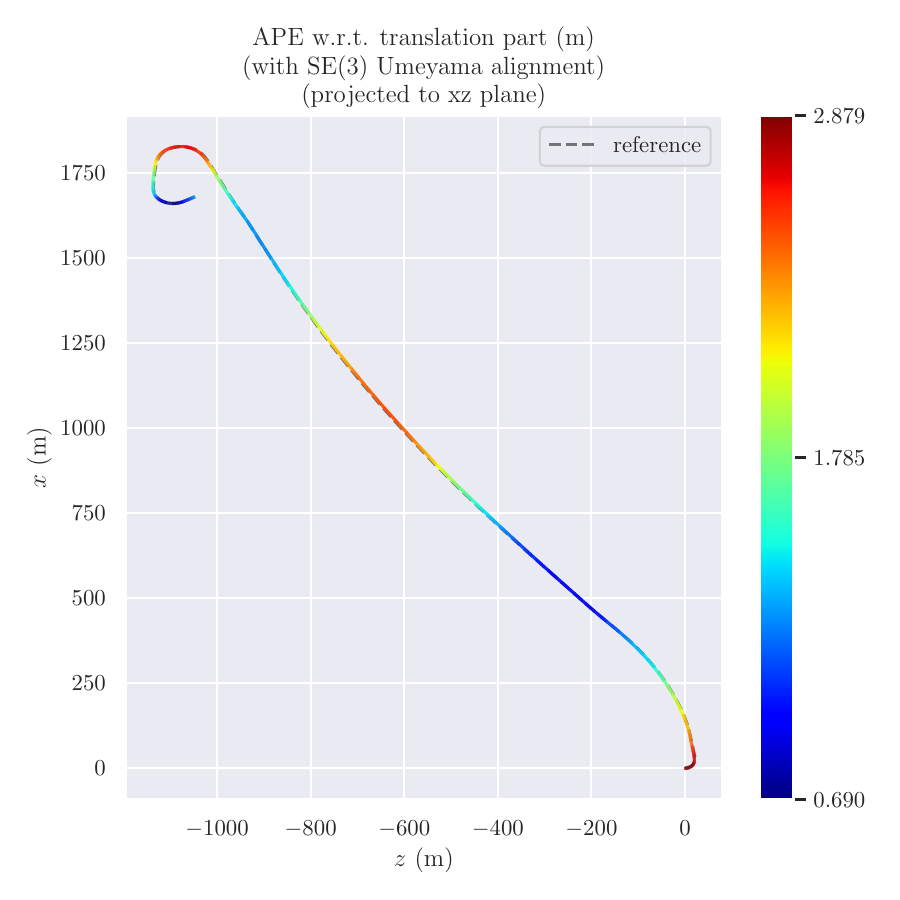}
        \caption{Scenario 01 (2D)}
		\label{fig:01-2d}
    \end{subfigure}\hfill
    \begin{subfigure}[b]{0.245\textwidth}
		\includegraphics[width=\textwidth]{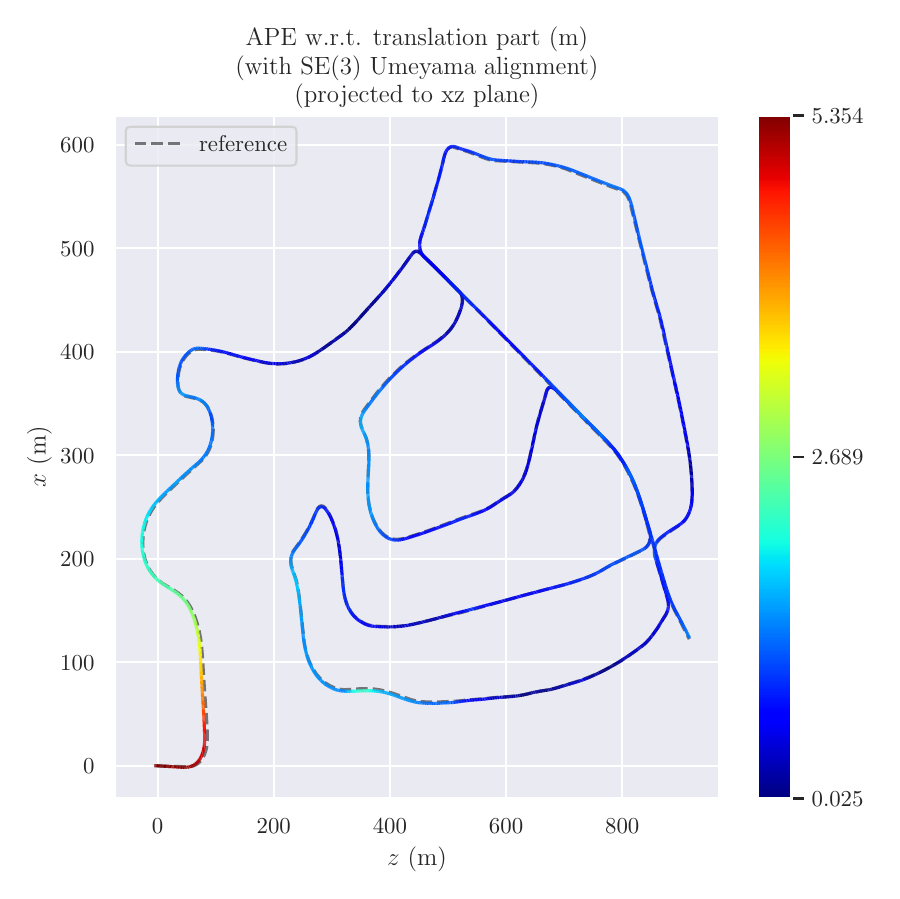}
        \caption{Scenario 02 (2D)}
		\label{fig:02-sd}
    \end{subfigure}\hfill
    \begin{subfigure}[b]{0.245\textwidth}
		\includegraphics[width=\textwidth]{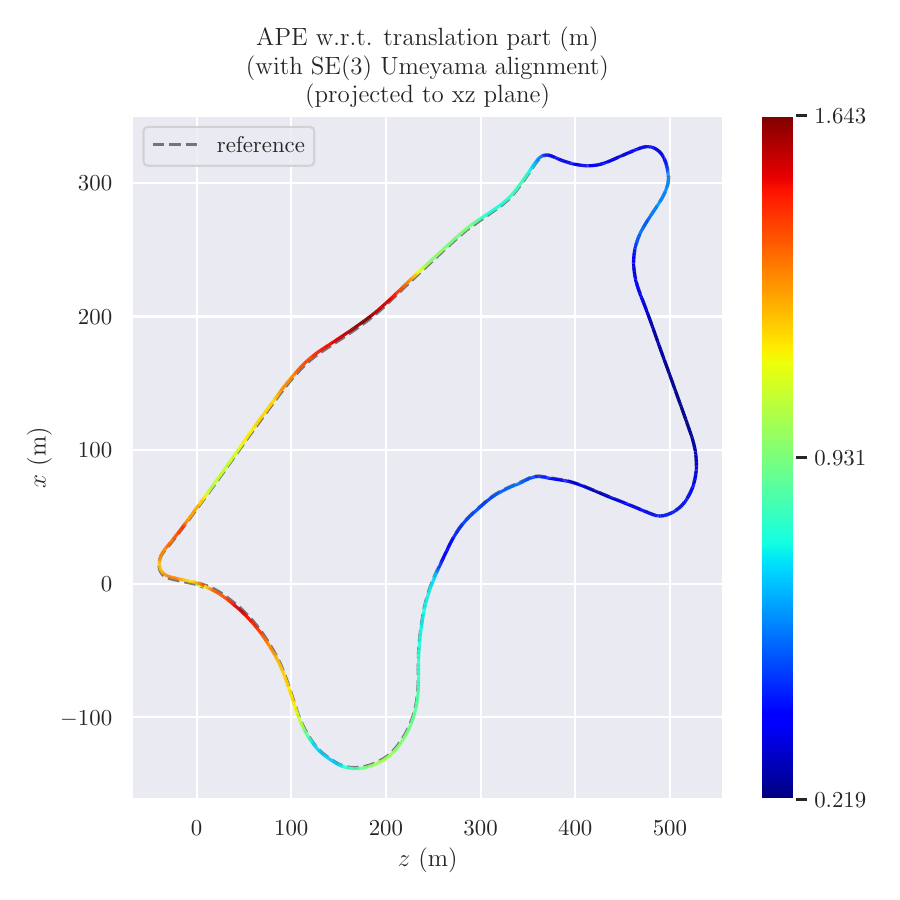}
        \caption{Scenario 09 (2D)}
		\label{fig:09-2d}
    \end{subfigure}\hfill
    \caption{Exmaple trajectories of the KITTI odometry dataset. The graphics are generated utilizing \cite{grupp_evo_2017}.}
    \label{fig:kitti_odometry_trajectories}
\end{figure*}

In this section, we present the methodology employed in this work. The foundation of our approach is KISS-ICP, known for its capabilities in achieving accurate LiDAR odometry. This is integrated with the loop closure functionality of Cartographer. Cartographer's scan matching process relies on a Ceres-based framework \cite{agarwal_ceres_2023}, which is a primary source of inaccuracies in large-scale environments. This limitation is evident in \cref{tab:rte}, \cref{tab:ate3d} and \cref{tab:ate2d}, where Cartographer is compared with related LiDAR odometry approaches. To address this issue, the odometry module is replaced with the state-of-the-art KISS-ICP \cite{vizzo_kiss-icp_2023}.

Our contribution extends beyond this integration by adapting KISS-ICP to handle semantically labeled point clouds. Modifications include changes to the sub sampling process, where dynamic obstacles are first filtered out from the original point cloud. In its original implementation, KISS-ICP performs point cloud sub sampling using a voxel grid of size $v\times v\times v$, where each cell can hold a given maximum number of points. We extend this approach to prioritize critical points based on their labels. Specifically, if a point's label corresponds to a critical object, such as poles, traffic signs, or traffic lights, it is retained even if the maximum number of points in a voxel is exceeded. This ensures that small or thin, but important, objects are preserved in the resulting map.

Additionally, KISS-ICP applies a secondary down sampling step to the already subsampled point cloud during the registration process. To preserve the representation of small objects, we adapt the voxel grid to incorporate semantic labels associated with the points, ensuring critical objects are retained throughout the process. Furthermore, an additional adaptation was introduced at the optimization stage for semantic KISS-ICP. The original formulation was extended by incorporating an additional parameter, 
$\kappa$, which evaluates the label fitness at the established correspondences.

\begin{equation}
    \Delta T_{est,j} = \underset{T}{\argmin} \underset{(s,q) \in C(\tau_t)}{\sum} \rho \cdot \kappa \cdot (\lVert Ts - q \rVert_2)
\end{equation}

Here, $C(\tau_t)$ represents the set of nearest neighbor correspondences with distances less than 
$\tau_t$, and $\rho$ denotes the Geman-McClure robust kernel. The Geman-McClure kernel is an M-estimator with strong outlier rejection capabilities. The fitness function in our case is defined as followed:

\begin{equation}
    \kappa(\cdot ,\cdot) = \begin{cases}
                            P(y_s \vert u) &, \text{if}\ y_s = y_q   \\
                            1- P(y_s \vert u)&, \text{else}  
                        \end{cases}
\end{equation}

which is using the certainty of the predicted label to weight the residual. The results utilizing this approach can also be seen in \cref{tab:rte} and \cref{tab:ate3d}.

After enhancing odometry estimation by incorporating semantic information into the ICP process, the resulting data is processed within the Cartographer framework. Cartographer’s loop closure algorithm operates in two stages. In the first stage, odometry estimates are used to generate local submaps, typically spanning approximately one second of incoming data. In our approach, semantic information is integrated during the submap creation process, extending the capabilities of the base implementation.

In the original Cartographer framework, ray-casting is used to create occupancy grids. LiDAR measurements determine the endpoints of rays, while intermediate cells along the rays are marked as unoccupied. This method requires filtering the ground plane, as ground points would otherwise dominate the occupancy grid and obstruct submap generation. Our approach leverages semantic information to overcome these limitations. Each measurement point is assigned to a multi-dimensional grid where each layer corresponds to a specific semantic label. The resulting submaps retain semantic information, as illustrated in \cref{fig:wd} and serve as inputs for the second stage of optimization.

In the second stage, submaps are aligned with one another. The original algorithm proposed by \cite{hess_real-time_2016} employs branch-and-bound scan matching to align occupancy grids. We adapt this algorithm to match semantic submaps instead. The alignment score is computed based on semantic labels, prioritizing the label with the highest hits within each grid cell rather than relying solely on the maximum value. This semantic-aware branch-and-bound approach enables efficient and robust loop closure detection.

Beyond eliminating the need for explicit ground segmentation, our method also supports post-processing of the resulting SLAM map. For instance, as shown in \cref{fig:wd_zoom} and \cref{fig:wd_zoom_no-car}, parked vehicles can be filtered out after the map is generated. This results in a cleaner and more accurate SLAM map that represents only static environments, improving its utility for re-localization tasks. By focusing on static features, our approach enhances both map reliability and the robustness of downstream applications.

\section{EVALUATION}
\label{sec:evaluation}

\begin{table*}[th]
\begin{tabularx}{\textwidth}{XXXXXXXXXXXXX}                                                                                               \multicolumn{13}{c}{Relative Trajectory Error (RTE) on the 11 scenarios of KITTI Odometry Benchmark}                                                                        \\ \hline
\multicolumn{1}{l|}{}    & 00   & 01   & 02   & 03   & 04   & 05   & 06   & 07   & 08   & 09   & 10    & Avg.      \\ \hline
\multicolumn{1}{l|}{LOAM}              & 0.78 & 1.43 & 0.92 & 0.86 & 0.71 & 0.57 & 0.65 & 0.63 & 1.12 & 0.77 & 0.79  & 0.84 \\
\multicolumn{1}{l|}{SA-LOAM}           & 0.59 & 1.89 & 0.79 & 0.87 & 0.59 & 0.37 & 0.52 & 0.41 & 0.84 & 0.77 & 0.78  & 0.76 \\
\multicolumn{1}{l|}{Suma++ (LC)}       & 0.65 & 1.63 & 3.54 & 0.67 & 0.34 & 0.40 & 0.47 & 0.39 & 1.01 & 0.58 & 0.67  & 0.94 \\
\multicolumn{1}{l|}{KISS-ICP}          & 0.51 & 0.72 & 0.52 & 0.66 & 0.35 & 0.30 & 0.26 & 0.32 & 0.82 & 0.49 & 0.57  & 0.50 \\
\multicolumn{1}{l|}{SAGE-ICP}          & 0.50 & 0.64 & \textbf{0.49} & 0.67 & \textbf{0.32} & \textbf{0.27} & \textbf{0.25} & 0.32 & 0.80 & 0.48 & 0.51  & \textbf{0.48} \\
\multicolumn{1}{l|}{SAGE-ICP$^{+}$}          & \textbf{0.47} & 1.48 & 1.10 & \textbf{0.43} & 0.53 & 0.38 & 0.44 & \textbf{0.27 }& \textbf{0.71} & \textbf{0.47} & \textbf{0.42} & 0.60 \\
\multicolumn{1}{l|}{Cartographer}      & 0.80 & \textbf{0.62} & 0.95 & 2.40 & 0.50 & 0.53 & 0.34 & 0.41 & 0.97 & 0.88 & 0.84  & 0.84 \\
\multicolumn{1}{l|}{Cartographer (LC)} & 0.80 & \textbf{0.62} & 1.01 & 2.41 & 0.50 & 0.49 & 0.30 & 0.36 & 0.92 & 0.79 & 0.83  & 0.82 \\
\hline
\multicolumn{1}{l|}{Ours}              & 0.52 & 0.78 & 0.51 & 0.70 & 0.37 & 0.33 & 0.29 & 0.37 & 0.83 & 0.52 & \textbf{0.51}  & 0.52 \\
\multicolumn{1}{l|}{Ours (LC)}         & 0.79 & 0.66 & 0.64 & 2.43 & 0.35 & 0.53 & 0.34 & 0.46 & 1.00 & 0.84 & 0.93  & 0.81    
\end{tabularx}
\caption{KISS-ICP and SAGE-ICP demonstrate the highest performance in the RTE metric, whereas our approach achieves comparably competitive results. LC denotes approaches with loop closure. \\
$^{+}$: Replicated results of us using https://github.com/nesc-iv/sage-icp}
\label{tab:rte}
\end{table*}

\begin{table*}[th]
\begin{tabularx}{\textwidth}{XXXXXXXXXXXXX}                                                                                               \multicolumn{13}{c}{Absolute Trajectory Error 3D on the 11 scenarios of KITTI Odometry Benchmark}                                                                        \\ \hline
\multicolumn{1}{l|}{}                  & 00   & 01   & 02   & 03   & 04   & 05   & 06   & 07   & 08   & 09   & 10    & Avg.      \\ \hline
\multicolumn{1}{l|}{KISS-ICP}          & 3.40 & 3.83 & 5.70 & \textbf{0.50} & 0.26 & 1.46 & 0.42 & 0.34 & 2.27 & 1.22 & 0.73  & 1.83 \\
\multicolumn{1}{l|}{SAGE-ICP$^{+}$}        & 4.40 & 12.11 & 7.86 & 0.72 & 0.52 & 1.39 & 0.73 & 0.34 & 4.12 & 2.07 & 1.10 & 3.10 \\
\multicolumn{1}{l|}{Cartographer}      & 7.21 & 4.63 & 12.54 & 1.03 & 0.34 & 2.22 & 0.28 & 0.39 & 2.72 & 2.10 & 0.92  & 3.13 \\
\multicolumn{1}{l|}{Cartographer (LC)} & 1.39 & 4.62 & 6.66 & 1.03 & 0.34 & \textbf{1.09} & \textbf{0.22} & \textbf{0.28} & 2.14 & 1.45 & 0.92  & 1.83 \\
\hline
\multicolumn{1}{l|}{Ours}              & 3.23 & 3.70 & 5.10 & 0.51 & 0.23 & 1.41 & 0.41 & 0.35 & 2.18 & \textbf{1.17} & \textbf{0.66}  & 1.72 \\
\multicolumn{1}{l|}{Ours (LC)}         & \textbf{1.36} & \textbf{3.70} & \textbf{1.89} & 0.64 & \textbf{0.22} & 1.11 & 0.33 & 0.32 & \textbf{2.08} & 1.28 & 0.78  & \textbf{1.25}   
\end{tabularx}
\caption{Our method outperforms the current state-of-the-art approaches in the ATE-3D metric. On average our approach has an ATE of \SI{1.25}{\metre} while KISS-ICP only achieves \SI{1.83}{\metre}.  LC denotes approaches with loop closure.\\
$^{+}$: Replicated results of us using https://github.com/nesc-iv/sage-icp}
\label{tab:ate3d}
\end{table*}

\begin{table*}[th]
\begin{tabularx}{\textwidth}{XXXXXXXXXXXXX}                                                                                               \multicolumn{13}{c}{Absolute Trajectory Error 2D on the 11 scenarios of KITTI Odometry Benchmark}                                                                        \\ \hline
\multicolumn{1}{l|}{}                  & 00   & 01   & 02   & 03   & 04   & 05   & 06   & 07   & 08   & 09   & 10    & Avg.      \\ \hline
\multicolumn{1}{l|}{KISS-ICP}          & 3.18  &2.12  & 5.53 & \textbf{0.48} & 0.25 & 1.00 & 0.42 & 0.29 & 1.47 & 0.78 & 0.70 & 1.48 \\
\multicolumn{1}{l|}{SAGE-ICP$^{+}$}        & 3.05  &11.40 & 5.69 & 0.64 & 0.51 & 0.96 & \textbf{0.11} & 0.33 & 3.03 & 1.39 & 0.92 & 2.55 \\
\multicolumn{1}{l|}{Cartographer}      & 0.68  &3.33  & 6.53 & 1.02 & 0.32 & \textbf{0.23} & 0.21 & \textbf{0.21} & 1.27 & 1.12 & 0.90 & 1.44 \\
\multicolumn{1}{l|}{Cartographer (LC)} & 7.11  &3.34  &12.46 & 1.02 & 0.32 & 1.95 & 0.27 & 0.34 & 2.10 & 1.87 & 0.89 & 2.88 \\\hline
\multicolumn{1}{l|}{Ours}              & 3.00  &1.90  & 4.90 & 0.50 & 0.22 & 0.93 & 0.40 & 0.29 & 1.34 & \textbf{0.71} & \textbf{0.62} & 1.35 \\
\multicolumn{1}{l|}{Ours (LC)}         & \textbf{0.63}  &\textbf{1.90}  & \textbf{1.35} & 0.63 & \textbf{0.21} & 0.31 & 0.33 & 0.26 & \textbf{1.17} & 0.88 & 0.75 & \textbf{0.77} \\ 
\end{tabularx}
\caption{Our method outperforms the current state-of-the-art approaches in the ATE-2D metric. On average our approach has an ATE of \SI{0.77}{\metre} while KISS-ICP only achieves \SI{1.48}{\metre}.\\
LC denotes approaches with loop closure. \\
$^{+}$: Replicated results of us using https://github.com/nesc-iv/sage-icp}
\label{tab:ate2d}
\end{table*}

The KITTI Odometry Benchmark \cite{geiger_are_2012} comprises 11 training sequences that include LiDAR scans for evaluating odometry and mapping algorithms. Behley et al. enhance these sequences in the SemanticKITTI dataset \cite{behley_semantickitti_2019, behley_towards_2021} by providing semantic labels for all 11 training sequences, thereby enabling detailed analysis of semantic mapping techniques.

The first metric evaluates the relative trajectory error (RTE) of the KITTI odometry dataset. This metric emphasizes odometry accuracy. Starting from every pose on the trajectory, it compares the pose error of the start and end poses of segments from \SI{100}{\metre} to \SI{800}{\metre} in length. It is therefore especially susceptible to orientation errors.

Additionally, we compute the Absolute Trajectory Error (ATE), which assesses the root mean square error (RMSE) of pairwise distances between the estimated and ground-truth trajectories in both 3D and 2D spaces. Before calculating the ATE, the estimated trajectories are aligned to the ground-truth trajectories using the Umeyama algorithm, ensuring a fair comparison by eliminating global misalignment. The results of these evaluations are presented in \cref{tab:ate3d} and \cref{tab:ate2d}.

For benchmarking, we compare our proposed approach against the state-of-the-art SAGE-ICP method, which serves as the baseline for evaluation. Additional comparisons are performed with other prominent methods, including KISS-ICP, LOAM, SuMa++, SA-LOAM, and Cartographer. Among these, LOAM represents a purely LiDAR-based odometry and mapping approach for the KITTI dataset, whereas SuMa++ exemplifies semantic-aided LiDAR SLAM systems. Our analysis incorporates both the RTE and ATE metrics to provide a comprehensive assessment of performance, with an emphasis on accurate map generation. Furthermore, we distinguish between 2D and 3D ATE metrics to provide a nuanced understanding of the method's strengths in real-world map representation.

The odometry-specific evaluation metrics for the KITTI dataset, focusing on RTE, are summarized in \cref{tab:rte}. This metric isolates the performance of odometry estimation without considering the overall map reconstruction. Our approach achieves an average RTE of 0.52, which, while slightly higher than SAGE-ICP's 0.48, demonstrates competitive performance. It is important to note that SAGE-ICP requires fine-tuning for each dataset and employs a dataset-specific parameter set. In contrast, our method operates with a more generalized configuration, requiring no parameter adjustments for specific use cases. Despite this generalization, our odometry-only approach remains competitive with both KISS-ICP and SAGE-ICP. Compared to the other Loop-Closure approaches, the result of our approach delivers the best results with  0.81 compared to the 0.94 of Suma++ and 0.82 of Cartographer. 

The ATE evaluation, encompassing both 2D and 3D results, highlights the strengths of our approach in generating accurate environmental maps. As shown in \cref{tab:ate2d} and \cref{tab:ate3d}, our method significantly outperforms the state-of-the-art KISS-ICP in the ATE-3D evaluation, achieving an improvement of \SI{0.11}{\metre} or 6\% on average, when only odometry is evaluated. When loop closure is enabled, this performance further improves by \SI{0.58}{\metre} or 31.7\%. In the ATE-2D evaluation, our approach performs nearly twice as well compared to KISS-ICP, decreasing the error from \SI{1.48}{\metre} without loop closure to \SI{0.77}{\metre} with loop closure. In \cite{cui_sage-icp_2023}, the authors evaluated against the KITTI odometry benchmark using the RTE. We reevaluated their approach to also calculate the ATE, but failed to replicate their results in the RTE metric. These results underline the efficacy of our method in generating precise and consistent map representations that closely align with real-world environments.

This is further illustrated through the direct comparison in \cref{fig:kitti_odometry_trajectories}. The absolute error is notably reduced in the 2D space, indicating that the loop closure algorithm significantly enhances the 2D pose estimation. Moreover, the y-component in the KITTI odometry emerges as the primary contributor to the observed error.

Overall, our approach demonstrates a compelling combination of generalization, competitive odometry performance, and superior map accuracy, particularly in scenarios requiring comprehensive 3D and 2D environmental understanding.

\section{CONCLUSION}
\label{sec:conclusion}
This paper introduces an ICP algorithm that incorporates semantic information to enhance odometry estimation based on KISS-ICP, achieving performance competitive with state-of-the-art methods. Beyond this advancement, we integrate the proposed ICP approach into a loop closure framework, specifically Cartographer. This integration enables the creation of environment maps that are not only highly accurate but also suitable for re localization tasks.

The Cartographer framework is built as an independent system, relying on minimal external dependencies. This portability makes it adaptable for deployment on various test vehicles, offering significant flexibility for real-world applications. By leveraging semantic information and improving odometry estimation within the framework, our approach surpasses existing state-of-the-art methods in mapping accuracy.

For evaluation, we utilize the SemanticKITTI dataset as the ground truth. The performance of the proposed approach is assessed using two key metrics: RTE and ATE in both 2D and 3D. While our method demonstrates competitive performance in odometry estimation compared to state-of-the-art techniques, it achieves a substantial reduction in ATE. Specifically, the incorporation of our adapted loop closure framework within Cartographer reduces the absolute pose error by nearly 50\%.

This significant improvement highlights the robustness of our approach and its potential for future research, particularly in applications involving real-world data and large-scale mapping.

\section*{ACKNOWLEDGMENT}

This research was partially supported by the German Federal  Ministry  for  Digital and Transport  (BMDV) in  the  funding  program C2CBridge1 Country to City Bridge - Analyse und funktionale Konzepte, Karlsruhe (19DZ23003B).

\newpage

\printbibliography

\end{document}